\pdfoutput=1

\documentclass[11pt]{article}

\usepackage[]{ACL2023}

\usepackage{times}
\usepackage{latexsym}

\usepackage[T1]{fontenc}

\usepackage[utf8]{inputenc}

\usepackage{microtype}

\usepackage{inconsolata}
\usepackage{amssymb}
\usepackage{amsmath}
\usepackage{multirow}
\usepackage[normalem]{ulem}
\useunder{\uline}{\ul}{}
\usepackage{stfloats}
\usepackage{graphicx}
 \usepackage{booktabs} 

\usepackage{booktabs}

\title{``In Dialogues We Learn'': Towards Personalized Dialogue\\ Without Pre-defined Profiles through In-Dialogue Learning}

\author{
\begin{tabular}{c}
Chuanqi Cheng$^{1}$\thanks{\ \ Equal contribution.} \quad  Quan Tu$^{1}$\footnotemark[1] \quad \textbf{Shuo Shang}$^{3}$ \quad \textbf{Cunli Mao}$^{4}$ \\ \textbf{Zhengtao Yu}$^{4}$ $^\dag$ \quad \textbf{Wei Wu}$^{2}$ $^\dag$ \quad \textbf{Rui Yan}$^{1,5}$ \thanks{\ \ Corresponding authors.} 
\end{tabular}
\\ \vspace{.5mm}
    \small
    \begin{tabular}{c}
    $^1$Gaoling School of Artificial Intelligence, Renmin University of China \quad $^2$Ant Group \\ \quad $^3$University of Electronic Science and Technology of China \quad $^4$Kunming University of Science and Technology \\ \quad $^5$Engineering Research Center of Next-Generation Intelligent Search and Recommendation, Ministry of Education
    \end{tabular}
    \\ \vspace{.5mm}
    \small
    \begin{tabular}{c}
    \texttt{\{chengchuanqi,quantu,ruiyan\}@ruc.edu.cn}, \texttt{\{congyue.ww\}@antgroup.com}  \\
    \end{tabular}
    \\ \vspace{.5mm}
    \small
    \begin{tabular}{c}
    \texttt{\{jedi.shang\}@gmail.com}, \texttt{\{maocunli\}@163.com}, \texttt{\{ztyu\}@hotmail.com} \\
    \end{tabular}
    \vspace{2mm} \\
}

\begin{document}
\maketitle
\begin{abstract}
Personalized dialogue systems have gained significant attention in recent years for their ability to generate responses in alignment with different personas. However, most existing approaches rely on pre-defined personal profiles, which are not only time-consuming and labor-intensive to create but also lack flexibility. 
We propose \textbf{I}n-\textbf{D}ialogue \textbf{L}earning (IDL), a fine-tuning framework that enhances the ability of pre-trained large language models to leverage dialogue history to characterize persona for personalized dialogue generation tasks without pre-defined profiles. Our experiments on three datasets demonstrate that IDL brings substantial improvements, with BLEU and ROUGE scores increasing by up to $200\%$ and $247\%$, respectively. Additionally, the results of human evaluations further validate the efficacy of our proposed method.
\end{abstract}

\section{Introduction}

Recently, there has been growing interest in personalized dialogue systems~\cite{tang2023enhancing, chen2023learning, huang2023personalized, chen-etal-2023-towards-robust, tu-etal-2022-misc}. Such systems are often adept at incorporating special personal characteristics into responses. Consequently, they offer enhanced flexibility, enabling them to adapt more effectively to a wide range of conversational scenarios, such as 
personal assistants or chatbots~\footnote{https://character.ai/}.

\begin{figure}[!h]
\centerline{\includegraphics[width=0.5\textwidth]{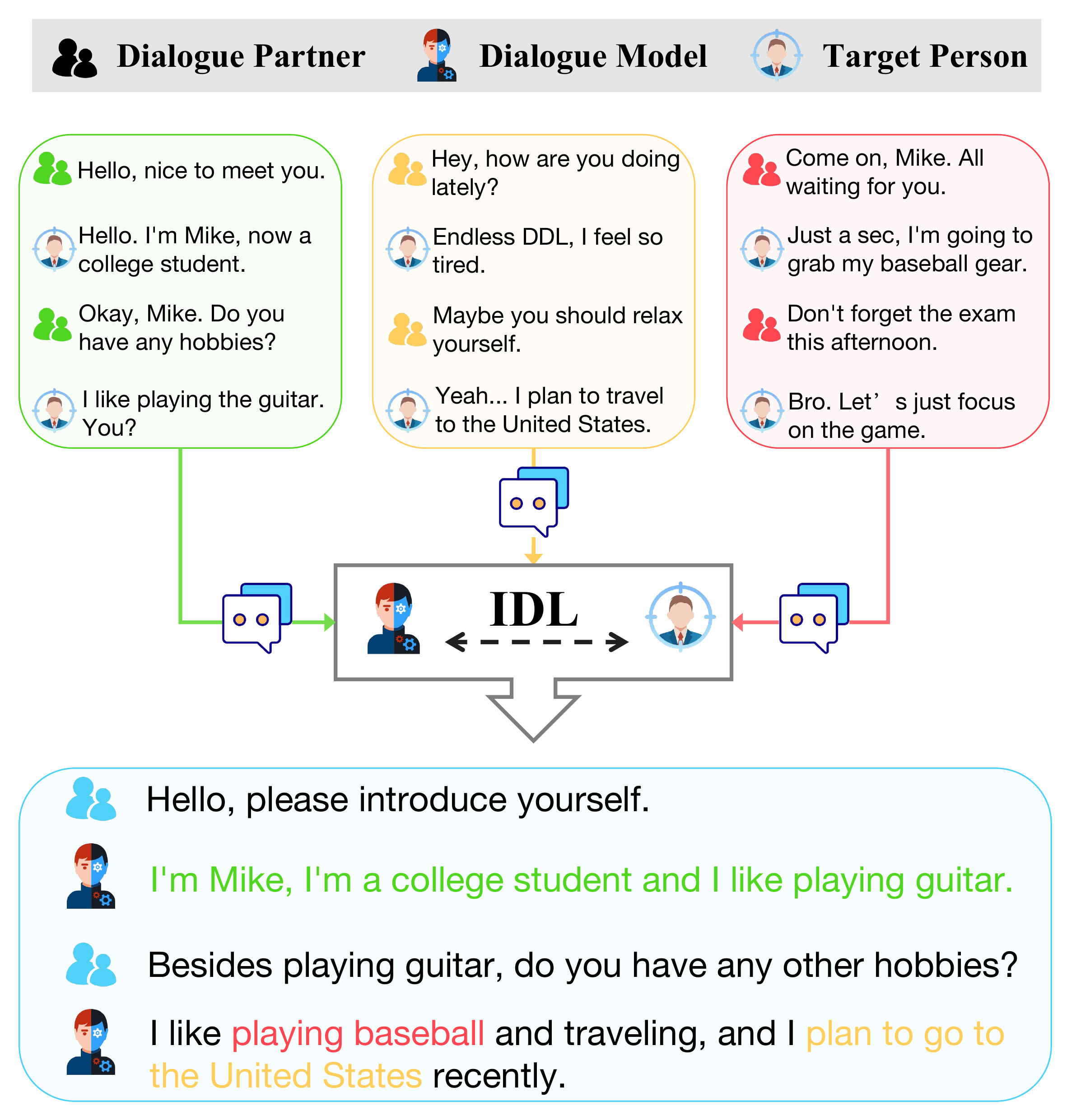}}
\caption{An example of profile-free personalized dialogue generation by In-Dialogue Learning. Persona information in different dialogues is marked with corresponding colors.}
\label{fig:profile-free}
\end{figure}
A common practice in personalized dialogues is to condition a dialogue model on a pre-defined profile that explicitly depicts the personality traits one aims to portray with a textual description. While there have been extensive studies along this line~\cite{song2021bob, liu2022improving, chen2023towards}, we explore the problem from a different angle: instead of using a brief profile to describe a person's personality, we leverage multiple conversations between the individual and others to build a personalized dialogue model. Consequently, the model can generate personalized dialogues without the need for pre-designed profiles which could be both time-consuming and labor-intensive. Furthermore, as dialogue history accumulates, past conversations may provide more personalized information than a static profile.



We introduce \textbf{I}n-\textbf{D}ialogue \textbf{L}earning (IDL), a two-stage framework that directly learns persona information from dialogue sessions, and leverages the learnt insights to synthesize responses that exhibit explicit personality characteristics (cf., Figure~\ref{fig:profile-free}). IDL comprises a Mutual Supervised Learning (MSL) stage and a Deep Personalized Alignment (DPA) stage. The objective of MSL is to equip a dialogue model with persona knowledge conveyed in dialogue sessions. To this end, one can simply select one dialogue as the target and take the remaining as the reference to perform few-shot learning to optimize the dialogue model. Such a straightforward implementation, however, suffers from two major problems: (1) unified reference dialogues normally contain abundant irrelevant information to the target dialogue, which increases the difficulty of learning; and (2) incoherent transition in multiple dialogues could cause disruption in the dialogue structure. 
To address the problems, we propose Static Persona Identification (SPI) and Dynamic Persona Identification (DPI) to cluster and re-order dyadic dialogues between a target person and the other interlocutors for effective IDL. 
SPI divides the dialogues of the person into multiple persona-relevant clusters, ensuring that the target dialogue can easily access inter-session personalized information from reference dialogues from each cluster.
DPI further re-orders the reference dialogues by minimizing the gaps 
in these dialogues, which is measured by conversational edit distance (convED)~\cite{lavi2021we}.


To better align responses with the target persona~\cite{ouyang2022training, yuan2023rrhf, song2023preference, hong2023cyclealign}, we introduce Direct Preference Optimization with Criterion (DPOC), an optimization method derived from DPO~\cite{rafailov2023direct} to mitigate preference degradation problem with a criterion-based penalty. This approach ensures that responses are more closely aligned with the target persona learned from reference dialogues.

We conduct experiments on several personalized dialogue datasets to evaluate the effectiveness of IDL.
Evaluation results show that IDL achieves performance comparable to very strong profile-based methods, without utilizing any pre-defined profile information.  
In comparison to traditional personalized dialogue approaches, IDL demonstrates significant improvements, highlighting the benefits of leveraging large language models for personalized dialogue. Furthermore, IDL shows significant improvement over In-Context Learning (ICL) when both utilize large language models, with BLEU and ROUGE scores increasing up to $200\%$ and $247\%$, respectively. This suggests that, unlike ICL, which primarily learns from data samples (single-turn), IDL is more effective at incorporating persona information within dialogues (multi-turn).

Our contributions are threefold:

\ \ (1) We introduce In-Dialogue Learning (IDL) as the first effort to create a personalized dialogue system using large language models without pre-defined user profiles, enabling response generation using persona information directly learned from dialogue sessions.

\ \ (2) We introduce methods for static and dynamic persona identification to improve data organization for IDL and enhance the use of persona information from dialogues. Additionally, we present DPOC, a novel reinforcement learning approach, to address preference degradation problem and align responses more precisely with the persona indicated in reference dialogues.

\ \ (3) We conduct extensive experiments on multiple datasets, showing the superior performance of IDL on personalized dialogue generation. As a profile-free method, it achieves comparable performance with profile-based methods and significantly outperforms other profile-free methods.

\section{Related Work}
\subsection{Personalized Dialogue Systems}
Personalized dialogue methods are classified into three types based on persona information acquisition. The first type uses structured databases (e.g., tables) ~\cite{zhang2018personalizing, song2019exploiting, wolf2019transfertransfo, liu2020you, bao2019plato, song2021bob} but faces limitations in response diversity due to data sparsity. The second type uses plain text profiles for richer information~\cite{qian2018assigning, song2020profile, zheng2020pre, song2021bob, tang2023enhancing}, yet struggles to completely capture personality and requires significant effort, affecting scalability.

Different from these methods, the third type mines persona information from dialogue sessions. For example,
DHAP~\cite{ma2021one} uses a transformer-based approach to analyze dialogue history for generating responses, but it ignores partner utterances, missing key persona details.
MSP~\cite{zhong2022less} improves upon DHAP by using a retrieval method to collect similar dialogues from various users, yet it only selects limited tokens from these dialogues, affecting their coherence.
Our method, in a broad sense, belongs to the third type. The stark difference is that we make good use of the capabilities of large language models, and significantly enhance the performance of personalized dialogue systems when no profiles are available.

\subsection{In-Context Learning}
In-context learning (ICL) emerges as language models scale~\cite{ brown2020language, chowdhery2023palm, touvron2023llama}, enabling them to perform complex tasks by learning from a few contextual demonstrations~\cite{wei2022chain}. The ICL ability of LLMs can be enhanced by using supervised fine-tuning methods, involving in-context data construction and multitask learning~\cite{chen2022improving, min2021metaicl}, since pre-training objectives aren't designed for ICL. Researches also show that the effectiveness of ICL relies on the choice and arrangement of demonstrations~\cite{zhao2021calibrate, lu2021fantastically, chen-etal-2023-towards-robust}.

Our method, while looks similar to ICL, is tailored for personalized dialogue generation by organizing sessions and learning persona-related information, differing from typical supervised in-context fine-tuning. It also uniquely incorporates reinforcement learning to enhance personalized dialogue capabilities beyond ICL methods.

\section{Method}

\begin{figure*}[!h]
\centerline{\includegraphics[width=1.0\textwidth]{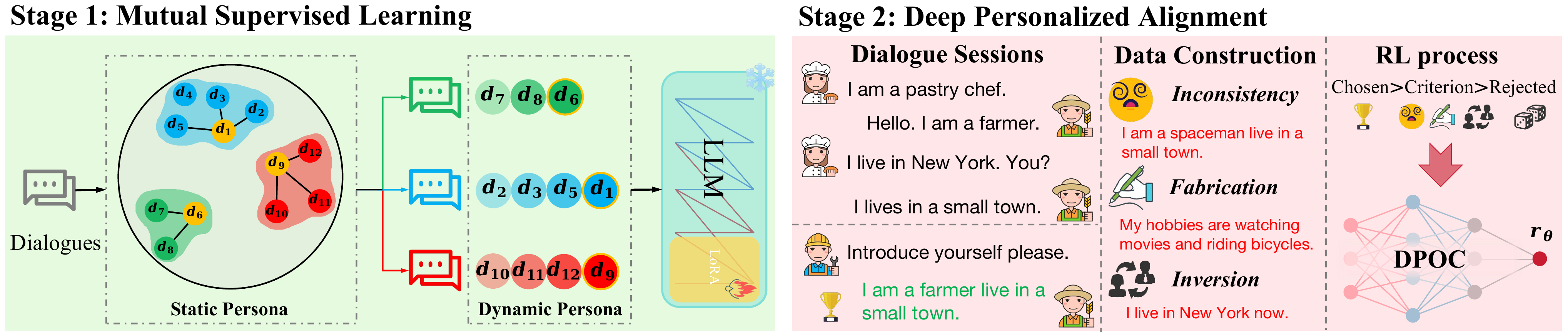}}
\caption{The framework of IDL. \textbf{Left}: the MSL stage that fine-tunes the dialogue model using data organized by static persona and dynamic persona identification. \textbf{Right}: the DPA stage in which we collect three types of criterion examples and conduct DPOC to further optimize the model to align with the target persona in a better way.}
\label{fig:overview}
\end{figure*}

As shown in Figure~\ref{fig:overview}, In-Dialogue Learning (IDL) involves two stages: Mutual Supervised Learning (MSL) and Deep Personalized Alignment (DPA). 
In the MSL stage, we propose static and dynamic persona identification to cluster and re-order the dialogues of the target person, and then organize these dialogues into an end-to-end form to perform supervised learning, endowing the model with the ability to leverage persona information within previous dialogues.
In the DPA stage, we further extend the DPO algorithm with Criterion (abbreviated as DPOC) to address the issue of preference degradation through the incorporation of criterion examples and penalty terms, facilitating fine-grained personalized learning.

\subsection{Problem Formalization}
\label{sec:problem formalization}

The goal of IDL is to generate responses that reflect the personality of a target person $u$ based on his/her previous dialogues $\mathbb{D}^u$. 
Formally, $\forall d^{(u,v)} = (q_1, r_1, \ldots, q_t, r_t) \in \mathbb{D}^u$, $d^{(u,v)}$ represents a dialogue between user $u$ and the other participant $v$ where $(q_i, r_i)$ is the $i$-th turn with $q_i$ the utterance from $v$ and $r_i$ the response from $u$, respectively. 
Given the current dialogue context $C_i=(q_1, r_1, \dots, q_i)$, the generation of IDL can be formulated as 
\begin{equation}
\label{eq:formula}
\small
    r_i = \text{LM}_{\Theta}(C_i, \mathbb{D}^u),
\end{equation}
where LM represents the language model, and $\Theta$ is the learnable parameters. Following the common practice, we concatenate $\mathbb{D}^u$ and $C_i$ as the input.

\subsection{Mutual Supervised Learning}
\label{sec:MSL}
IDL represents learning the personalized response generation ability conditioned on the previous dialogues. 
If we deem the dialogues of the target person as nodes in a graph, each of them can utilize the remaining dialogues as the reference, which can be imagined as a \textit{complete graph}. 
This property induces the concept of Mutual Supervised Learning (MSL).
However, the straightforward \textit{complete graph} usage suffers from two challenges: (1) over messy historical information and (2) incoherent transition relationship. 
The former denotes that the messy historical information will cause the misuse of persona information 
when dialogues with unrelated persona knowledge are used as the reference. 
The latter means that the improper order of these dialogues as the reference will cause incoherent cross-dialogue transition, harming the dialogue structure.
To overcome these two challenges, we propose \textbf{static and
dynamic persona identification} for personalized dialogue clustering and re-ordering (as shown in the left part of  Figure~\ref{fig:overview}). 

\subsubsection{Static Persona Identification}
\label{sec:spi}
Learning dialogue generation from a wide variety of reference dialogues is not always effective~\cite{bao2019plato}, especially when we aim to capture the personality characteristics embedded in the dialogues. To enhance the efficacy of the process, static persona identification partitions the dialogues of a target person into multiple persona-relevant clusters (cf., Figure~\ref{fig:overview} left). Hence, within each persona-relevant cluster, IDL can learn more meaningful mapping from reference dialogues to target dialogues. The challenge then lies in how to measure the distance between the dialogues across persona dimensions for effective dialogue clustering.

We employ a public dataset PersonaExt~\cite{zhu2023paed} and train a persona extractor to recognize persona-intensive utterances in a dialogue corpus. 
PersonaExt segregates persona information within dialogues into triples of \textit{<subject, relationship, object>}. 
The dataset defines $105$ types of relationships. 
Based on the dataset, we develop the persona extractor (abbreviated as Ext) that can directly extract the above-mentioned triples from a dialogue by fine-tuning an LLM. 
More details about training of the persona extractor are presented in Appendix~\ref{appendix: details}.
The persona extractor identifies and extracts persona-intensive utterances from a dialogue by recognizing utterances that contain at least one object in the extracted triples. Formally, the extraction process can be formulated as
\begin{equation}
\small
    \{p_j^{(u,v)}\}_{j=1}^{n} = \text{Ext}(d^{(u,v)}),
\end{equation}
where $p_j^{(u,v)}$ is a persona-intensive utterance in dialogue $d^{(u,v)}$, with $u$ and $v$ as the participants. $\{p_j^{(u,v)}\}_{j=1}^{n}$
 are then encoded as the persona representation $z^{(u,v)}$ of $d^{(u,v)}$ by
\begin{equation}
\small
\begin{aligned}
    p^{(u,v)} &= \text{Concat}(p_{1}^{(u,v)}, \dots, p_{n}^{(u,v)}),\\
 z^{(u, v)} &= \text{Enc}(p^{(u,v)}),
\end{aligned}
\end{equation}
where Enc$(\cdot)$ is an off-the-shelf sentence encoder\footnote{https://huggingface.co/sentence-transformers/all-mpnet-base-v2}.
Based on $\{z^{(u,v)}\}$, dialogues $\{d^{(u,v)}|d^{(u,v)} \in \mathbb{D}^u\}$ are clustered by k-means algorithm\footnote{We tested various clustering algorithms, including k-means, BSCAN, Mean Shift, WARD, and BIRCH, but observed no significant differences. Therefore, we chose k-means for its simplicity and widespread use.}:
\begin{equation}
\small
K^u = \text{KMeans}(\{z^{(u,v)}\}, c),
\end{equation}
where $z^{(u,v)}$ serves as the index of dialogue $d^{(u,v)}$ and $c$ is the number of clusters. 
Subsequently, within each cluster $K_j^u \in K^u, j=1,2,\dots,c$,  we randomly select a dialogue as the \textit{target dialogue} while the closest top-$k$ in the remaining dialogues are regarded as the \textit{reference dialogues}. 

\subsubsection{Dynamic Persona Identification}

Following static persona identification, we gather persona-relevant reference dialogues along with a target dialogue for optimization within each cluster. 
While we could directly concatenate these reference dialogues as input for the model, determining the optimal sequence remains a challenge. 
Our goal is to merge these dialogues into a cohesive long-term conversation, as we recognize that an inappropriate sequence could negatively affect the structure of the dialogue~\cite{chen2023towards}.

To achieve the goal, we compute the optimal order which could minimize the overall semantic distance between adjacent dialogue sessions in the long-term conversation. 
This approach ensures a smoother transition in the ongoing dialogue. 

To quantify the semantic distance between dialogues, we introduce Conversation Edit Distance (convED)~\cite{lavi2021we}. 
It is akin to the traditional edit distance, but it modifies the basic unit of editing from characters to sentences within a dialogue. 
The metric aligns one dialogue with another through the processes of inserting, deleting, and substituting sentences. Detailed formulations of convED are presented in Appendix \ref{appendix: convED}.

Given a pair of dialogues $(d_i, d_j)$, the distance $dist_{i,j} = \text{convED}(d_i, d_j)$ measures the cost of aligning $d_i$ to $d_j$. Hence, by computing paired convED, we obtain a semantic distance matrix between reference dialogues in a cluster. 
Subsequently, we introduce Dijkstra's minimum distance algorithm~\cite{dijkstra2022note} to re-order the reference dialogues based on the semantic distance matrix and compute the optimal order. 

In each cluster of $K^u$, we concatenate the reference dialogues according to the optimal order and split the target dialogue with the last utterance as a response and the remaining as the context. These data elements satisfy Equation~\ref{eq:formula}, and we can optimize the LM by minimizing the negative likelihood loss. The above processes endow the model with basic IDL ability, which could generate personalized responses based on reference dialogues. 

Note that we utilize two kinds of distance in static and dynamic persona identification, where the former measures the personalized relevance and clusters the relevant dialogues of a target person, while the latter measures the semantic distance and re-orders the reference dialogues in a cluster. 

\subsection{Deep Personalized Alignment}
The model after MSL initially exhibits the ability of personalized response generation by referencing some dialogues. However, due to hallucinations of LLMs~\cite{kalai2023calibrated}, it still falls short in generating more precise personalized responses.
Consequently, we propose Deep Personalized Alignment (DPA) in IDL.  

\subsubsection{DPOC}

Previous DPO~\cite{rafailov2023direct} method for preference alignment encounters a challenge in the form of unstable training outcomes. This instability is aroused by the primary objective of DPO, which maximizes the generation probability gap between chosen and rejected examples. This objective will overlook the diminishing rewards of the chosen examples.  
Thus, even when the disparity between chosen and rejected examples increases, it may be caused by a concurrent decrease in rewards for both chosen and rejected examples, ultimately leading to a diminished efficacy of the optimized model. This issue is referred as \textbf{preference degradation}.

To address this problem, DPOC incorporates a corrective measure by adding a penalty term $\mathcal{P}$:
\begin{equation}
\small
    \mathcal{P}(r_w, r_l) = - \min \left( 0 ,  \log r_w - \log r_l \right), 
\end{equation}
where $r_w$ is the reward of the better sample $y_w$ and $r_l$ is the reward of the worse sample $y_l$. In most cases, $r_w > r_l$ and $\mathcal{P}(r_w, r_l) = 0$. However, when $r_l > r_w$, $\mathcal{P}(r_w, r_l)$ functions as the penalty term. This inclusion ensures that the optimized model does not significantly deviate from the initial model. Building upon the foundation of DPO, 
the loss function of DPOC is formulated as 
\begin{equation}
\small
\begin{aligned}
    \mathcal{L}_{DPOC}(r_{cho}, r_{rej}, r_{crt}) &= \mathcal{L}_{DPO}(r_{cho}, r_{rej}) \\
    &+ \mathcal{P}(r_{cho}, r_{crt}) \\
    &+ \mathcal{P}(r_{crt}, r_{rej})
\end{aligned}
\end{equation}

The criterion sample reward $r_{crt}$ typically serves as intermediary pivots between chosen sample reward $r_{cho}$ and rejected sample reward $r_{rej}$. 
Specifically, if the reward from a chosen sample falls below that of a criterion sample, or if the reward of a rejected sample's reward is unexpectedly high compared to criterion examples, the current model incurs a penalty, which is represented by $\mathcal{P}(r_{cho}, r_{crt})$ and $\mathcal{P}(r_{crt}, r_{rej})$, respectively. Detailed formulations are presented in Appendix~\ref{appendix: dpoc}.

\subsubsection{Data Construction}
\label{sec: data_construction}
To perform DPOC, we need to specify the criterion samples. The intuition of criterion sample construction comes from analysis of the model after the MSL stage, where we observe three major problems, including responses revealing fictitious persona information, conflicting with the persona set by the context, and confusing the partner's persona with the target person's. Based on the analysis, we consider the following three types of criterion samples (cf., Figure~\ref{fig:overview} right):
(1) Inconsistency: includes information conflicting with the persona established in the dialogue sessions. (2) Fabrication: introduces personality details not mentioned in the dialogue sessions. (3) Inversion: adopts the persona information of the other participant. 

Given dialogue sessions $\mathbb{D}^u$, the context of ongoing dialogue $C$ and a chosen sample $h_{cho}$ of the current response, the construction of the three types of criterion examples are detailed as follows:

\noindent \textbf{Inconsistency.} We employ the personality extraction model introduced in $\S$\ref{sec:spi}, and
utilize the personality triplet randomly extracted from $\mathbb{D}^u$ to substitute a triplet in $h_{cho}$ to formulate $h_{crt}$. For example, $h_{cho}$ \textit{``I am a farmer live in a small town''} is transformed into $h_{crt}$ \textit{``I am a spaceman live in a small town''} by replacing  \textit{<I, job, farmer>} with \textit{<I, job, spaceman>}, which is extracted from $\mathbb{D}^u$.

\noindent \textbf{Fabrication.}
We encode sentences in the dataset, selecting  top-$m$ candidates with the highest semantic similarity to $h_{cho}$. A candidate, $h_{crt}$, is randomly chosen ensuring $\text{Ext}(h_{crt}) \cap \text{Ext}(\mathbb{D}^u) = \emptyset$. For example, from the utterance \textit{``My hobbies are watching movies and riding bicycles''}, we extract triples \textit{<I, hobby, watching movies>} and \textit{<I, hobby, riding bicycles>}. As the triples are not involved in $\text{Ext}(D^u)$, we can adopt this utterance as $h_{crt}$.

\noindent \textbf{Inversion.} 
In $\mathbb{D}^u$ and $C$, utterances are divided into $R$ for the target person $u$ and $Q$ for the other participant $v$, then the most semantically similar utterance in $Q$ to a chosen $r_{cho}$ is identified as $h_{crt}$. For instance, for $r_{cho}$ \textit{``I am a farmer living in a small town''}, \textit{``I live in New York''} from $Q$ is selected as $h_{crt}$.

\section{Experiments}

\subsection{Datasets}

\noindent \textbf{ConvAI2}~\cite{dinan2020second} is a high-quality English dataset focused on personalized dialogues. Each dialogue revolves around a specific profile. The dataset is expanded from the classic PersonaChat~\cite{zhang2018personalizing} by crowd workers. 
\textbf{Cornell Movie-Dialogs Corpus}~\cite{danescu2011chameleons} contains over $220,000$ dialogues collected from more than $600$ movies with rich meta-data, offering a diverse range of dialogues between $10,000$ pairs of characters.
\textbf{LIGHT}~\cite{urbanek2019learning} is a large-scale crowdsourced fantasy text adventure game research platform.
We extract dialoigues of each character to form the dataset used in the experiments.

Note that profiles are only available in ConvAI2 and not in Cornell Movie-Dialogs Corpus and LIGHT. Implementation details are presented in Appendix~\ref{appendix: details}.

\subsection{Baselines}
\textbf{Profile-based Approaches} utilize persona information extracted from the given profiles. Along this research line, we consider the following models:
{GPT-2}~\cite{radford2019language} is known for its proficiency in a variety of text generation tasks. 
{PerCVAE}~\cite{zhao2017learning} processes the persona information as a conditional representation and employs CVAE to produce personalized responses. 
{BoB}~\cite{song2021bob} leverages BERT for personalized dialogues by combining consistency generation task and consistency inference tasks.
{CLV}~\cite{tang2023enhancing} categorizes persona descriptions into distinct groups to enhance personalized response generation with historical queries.

\noindent \textbf{Profile-free Approaches} perform personalized dialogue generation without profiles. We employ \textit{DHAP}~\cite{ma2021one} and \textit{MSP}~\cite{zhong2022less} as baselines.

\noindent \textbf{Large Language Models} have made great progress recently. We select LLaMA-2-7B-Chat and LLaMA-2-13B-Chat~\cite{touvron2023llama} as the backbones of IDL, and name the models LLaMA-2-7B IDL and LLaMA-2-13B IDL, respectively. 
Besides, Vicuna\footnote{https://lmsys.org/blog/2023-03-30-vicuna/} and WizardLM~\cite{xu2023wizardlm} are involved in comparison, where the former is an open-source chatbot developed by fine-tuning LLaMA with ShareGPT, and the latter is fine-tuned from LLaMA-2, starting with a basic set of instructions.



In the ConvAI2 dataset, we compare our IDL models with both profile-based and profile-free approaches. Unlike existing profile-based methods that don't use Large Language Models (LLMs), we fine-tune LLaMA-2 models (7B and 13B versions) with ConvAI2's profiles for a fair comparison, naming them LLaMA-2-7B gold and LLaMA-2-13B gold. We also include two other LLM baselines: LLaMA-2 System, which uses profiles directly in system instructions without further training, and LLaMA-2 FT, which fine-tunes on ConvAI2 treating each conversation as a separate example.

For the Movie and LIGHT datasets, we test the adaptability of our IDL models (LLaMA-2-7B IDL and LLaMA-2-13B IDL, both fine-tuned on ConvAI2) against other LLMs using ICL.





\subsection{Evaluation Metrics}
We employ BLEU~\cite{papineni2002bleu} and ROUGE-L~\cite{lin2004automatic} metrics to assess the coherence of the text.\footnote{We use NLTK to calculate both metrics.} For evaluating diversity, Distinct-1/2~\cite{li2015diversity, lv-etal-2023-dialogps} metrics are utilized. Additionally, P-F1~\cite{ma2021one}, P-Co (Persona Cosine Similarity)~\cite{zhong2022less} are used to measure persona consistency, while Con.Score, and Coh-Con.Score are used to measure the consistency between model responses and the given profiles in ConvAI2~\cite{tang2023enhancing}.


\subsection{Main Results}

\begin{table*}[t!]
\small
\centering
\begin{tabular}{clcccccc}
\hline
\toprule
\multirow{2}{*}{Dataset} & \multicolumn{1}{c}{\multirow{2}{*}{Model}} & \multicolumn{2}{c}{Coherence} & \multicolumn{2}{c}{Diversity} & \multicolumn{2}{c}{Persona} \\ 
\cmidrule(lr){3-4}\cmidrule(lr){5-6}\cmidrule(lr){7-8}
 & & BLEU-1 & ROUGE-L & Dist-1 & Dist-2 & Coh. & Coh-Con. \\ \midrule
\multirow{10}{*}{ConvAI2} & GPT-2 & 6.77 & 10.96 & 68.22 & 88.81 & 56.71 & 13.29 \\
 & PerCVAE & 6.89 & 10.54 & 67.48 & 89.46 & 53.26 & 12.95 \\
 & BoB & 7.85 & 12.46 & 63.85 & 85.02 & 62.47 & 15.97 \\
 & DHAP & 7.21 & 9.90 & 69.86 & 90.23 & 64.27 & 16.04 \\
 & MSP & 8.19 & 11.67 & 65.79 & 89.43 & 65.81 & 15.45 \\
 & CLV & 11.85 & 15.10 & 71.24 & 92.89 & 71.72 & \textbf{23.01} \\ 
 \cmidrule(lr){2-8}
 & LLaMA2-7B System & 7.22 & 9.56 & 72.21 & 94.39 & 98.87 & \textbf{22.32} \\
 & LLaMA2-7B FT & 50.23 & 18.04 & \textbf{88.32} & \textbf{97.45} & {\ul 97.41} & 8.63 \\
 & LLaMA2-7B IDL & {\ul 52.40}$^\dag$ & {\ul 18.98}$^\dag$ & {86.13}$^\dag$ & {96.97}$^\dag$ & {96.86}$^\dag$ & {13.26}$^\dag$ \\
 & LLaMA2-7B gold & \textbf{54.56} & \textbf{20.98} & {\ul 87.02} & {\ul 97.33} & \textbf{98.15} & {\ul 18.72} \\ 
 \cmidrule(lr){2-8} 
 & LLaMA2-7B System & 11.80 & 10.39 & 76.46 & 94.88 & 98.92 & {\ul 19.30} \\
 & LLaMA2-7B FT & 51.80 & 18.14 & {\ul 88.29} & \textbf{97.80} & 97.64 & 9.71 \\
 & LLaMA2-13B IDL & {\ul 54.48}$^\dag$ & {\ul 20.05}$^\dag$ & {87.78}$^\dag$ & {97.45}$^\dag$ & \textbf{98.48}$^\dag$ & \textbf{19.63}$^\dag$ \\
 & LLaMA2-13B gold & \textbf{55.32} & \textbf{21.58} & \textbf{88.49} & {\ul 97.78} & {\ul 98.10} & 17.77 \\ \bottomrule
\end{tabular}
\caption{Automatic evaluation on ConvAI2. All models are trained on this dataset. The best results are in \textbf{bold} and the second best results are \underline{underlined}. ``$\dag$'' indicates that our model passed
t-test with $p$-value $< 0.05$ in comparison to the best baseline. Results on BLEU-2/3/4 are presented in \ref{appendix:supplement_exp}.}
\label{table: table1}
\end{table*}
\begin{table*}[t!]
\small
\centering
\begin{tabular}{cclccccccc}
\toprule
\multirow{2}{*}{Dataset} & \multirow{2}{*}{Size} & \multicolumn{1}{c}{\multirow{2}{*}{Model}} & \multicolumn{3}{c}{Coherence} & \multicolumn{2}{c}{Diversity} & \multicolumn{2}{c}{Persona} \\ \cmidrule(lr){4-6}\cmidrule(lr){7-8}\cmidrule(lr){9-10} 
 &  &  & BLEU-1 & BLEU-2 & ROUGE-L & Dist-1 & Dist-2 & P-F1 & P-Co \\ \midrule
\multirow{7}{*}{Movie} & \multirow{3}{*}{7B} & Vicuna & {\ul 14.76} & {\ul 5.53} & 5.44 & {\ul 71.45} & 63.58 & 11.13 & 17.05 \\
 &  & LLaMA-2 ICL & 6.12 & 3.07 & {\ul 5.95} & 65.38 & {\ul 91.10} & {\ul 11.70} & {\ul 18.95} \\
 &  & LLaMA-2 IDL & \textbf{31.60}$^\dag$ & \textbf{11.74}$^\dag$ & \textbf{10.86}$^\dag$ & \textbf{89.86}$^\dag$ & \textbf{95.81}$^\dag$ & \textbf{19.95}$^\dag$ & \textbf{21.07}$^\dag$ \\ \cmidrule(lr){2-10} 
 & \multirow{4}{*}{13B} & Vicuna & 12.82 & 4.01 & 3.88 & 75.37 & 60.53 & 6.54 & 14.22 \\
 &  & WizardLM & {\ul 29.60} & {\ul 10.45} & {\ul 9.75} & {\ul 87.55} & {\ul 94.62} & {\ul 18.67} & {\ul 20.92} \\
 &  & LLaMA-2 ICL & 15.04 & 7.00 & 8.21 & 75.26 & 94.55 & 14.38 & 20.71 \\
 &  & LLaMA-2 IDL & \textbf{32.56}$^\dag$ & \textbf{13.00}$^\dag$ & \textbf{10.62} & \textbf{90.31}$^\dag$ & \textbf{97.24}$^\dag$ & \textbf{19.67} & \textbf{22.88} \\ \midrule
\multirow{7}{*}{LIGHT} & \multirow{3}{*}{7B} & Vicuna & {\ul 36.07} & {\ul 17.37} & {\ul 10.52} & {\ul 83.27} & 90.56 & 16.53 & 23.40 \\
 &  & LLaMA-2 ICL & 15.41 & 8.92 & 9.88 & 67.74 & {\ul 93.24} & {\ul 16.78} & \textbf{31.99} \\
 &  & LLaMA-2 IDL & \textbf{46.32}$^\dag$ & \textbf{22.01}$^\dag$ & \textbf{13.45}$^\dag$ & \textbf{83.90}$^\dag$ & \textbf{94.70}$^\dag$ & \textbf{20.18}$^\dag$ & {\ul 28.00}$^\dag$ \\ \cmidrule(lr){2-10} 
 & \multirow{4}{*}{13B} & Vicuna & 19.68 & 8.87 & 5.87 & 59.85 & 58.07 & 8.27 & 16.11 \\
 &  & WizardLM & {\ul 44.59} & {\ul 21.45} & {\ul 11.13} & {\ul 83.11} & {\ul 95.15} & {\ul 18.28} & 28.01 \\
 &  & LLaMA-2 ICL & 24.31 & 13.47 & 10.55 & 75.07 & 96.24 & 17.69 & \textbf{31.48} \\
 &  & LLaMA-2 IDL & \textbf{49.69}$^\dag$ & \textbf{24.64}$^\dag$ & \textbf{13.24} & \textbf{87.53}$^\dag$ & \textbf{97.54} & \textbf{20.28} & {\ul 30.95} \\ \bottomrule
\end{tabular}
\caption{Automatic evaluation on Movie and LIGHT. The best results are in \textbf{bold} and the second best results are \underline{underlined}. ``$\dag$'' indicates that our model passed
t-test with $p$-value $< 0.05$ in comparison to the best baseline. Results on BLEU-3 and BLEU-4 are presented in Appendix \ref{appendix:supplement_exp}.}
\label{table: table2}
\end{table*}
\subsubsection{Automatic Evaluation}
In Table~\ref{table: table1}, we compare the proposed method with existing personalized dialogue generation methods on ConvAI2. From the results, we can conclude that (1) when equipped with IDL, an open-source LLM can significantly outperform the existing methods in terms of almost all metrics, implying that IDL offers an effective way for leveraging LLMs in the task of personalized dialogue generation. (2) IDL can successfully discover persona information from dialogue sessions, comparing LLaMA-2 IDL with LLaMA-2 gold. Even without any hints from the profiles, IDL can still achieve comparable performance to the models fully supervised by the profiles.

In Table~\ref{table: table2}, we present results of IDL and other LLMs of comparable size on Movie and LIGHT. All the baseline models engage in personalized dialogue through ICL. Based on the results, we observe that (1) ICL underperforms in personalized dialogue generation, indicating that while ICL can handle the textual structure of dialogue sessions, it fails to effectively utilize persona information within these dialogues and (2) LLaMA-2-7B IDL and LLaMA-2-13B IDL fine-tuned on ConvAI2 also perform well on Movie and LIGHT. This confirms that the success of IDL is not due to the optimization for a particular dataset; rather, it stems from the ability to effectively utilize persona information in dialogues.

\subsubsection{Human Evaluation}
\begin{figure}[!h]
\centerline{\includegraphics[width=0.5\textwidth]{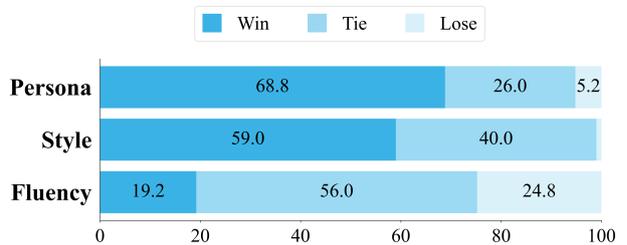}}
\caption{Human evaluation results for IDL compared to ICL. Both methods adopt LLaMA-2-13B-Chat.}
\label{fig:human}
\end{figure}
We hire $5$ well-educated volunteers as annotators, and require them to judge the quality of model responses from three aspects: \textbf{(1) Persona}: the annotators assess whether a response accurately and consistently reflects the persona information of the target person. \textbf{(2) Style}: the annotators judge if the response aligns with the expected wording and tone for the target person. \textbf{(3) Fluency}: the annotators examine the smoothness of the dialogue flow, considering both linguistic and logical fluency. We sample $500$ dialogues associated with demonstrations from the test set of ConvAI2, and obtain responses for each dialogue from IDL and ICL (based on LLaMA-2-13B-Chat), respectively. Each time, a pair of responses are randomly shuffled and presented to the $5$ annotators. Each annotator assign labels from \{Win, Tie, Lose\} to a pair according to Persona, Style, and Fluency, and in total, each pair obtains $5$ labels on each of the three aspects. Figure~\ref{fig:human} shows the evaluation results. IDL significantly improves upon ICL on both persona and style, with winning rates of 68.8\% and 59.0\%, respectively, demonstrating that the model using IDL can more effectively simulate the personality and tone of the target person. Regarding fluency, there is a slight decline in performance when using IDL, possibly attributed to the model's increased focus on aligning with persona information. We calculate Cohen's kappa, and the values for  persona, style, and fluency are $0.53$, $0.56$ and $0.51$, respectively, indicating moderate agreement among the annotators.

\subsection{Discussions}

\subsubsection{Ablation Study}
\begin{table}[!h]
\small
\centering
\begin{tabular}{lcccc}
\toprule
\textbf{Model} & \textbf{BLEU} & \textbf{ROUGE} & \textbf{P-F1} & \textbf{P-Co} \\ 
\midrule
IDL & \textbf{32.56} & \textbf{13.00} & \textbf{19.67} & \textbf{22.88} \\ 
\ \ \ \  \textit{w/o} Criterion & 31.58 & 10.55 & 17.76 & 21.79 \\
\ \ \ \  \textit{w/o} DPA & 31.25 & 10.89 & 18.98 & 21.12 \\
\ \ \ \  \textit{w/o} SPI & 29.94 & 10.93 & 19.02 & 21.14 \\
\ \ \ \  \textit{w/o} DPI & 28.80 & 9.60 & 18.46 & 21.01 \\ 
\bottomrule
\end{tabular}
\caption{Ablation study on Movie.}
\label{table: table3}
\end{table}
Table~\ref{table: table3} shows the ablation study results on Movie. In order to clarify the contribution of each IDL process to the overall effect, we gradually remove each process and get a list of variants: \textbf{(a)} \textit{w/o} Criterion removes the criterion samples and uses standard DPO for persona alignment. \textbf{(b)} \textit{w/o} DPA removes the whole persona alignment process. \textbf{(c)} \textit{w/o} SPI further removes the static persona identification in the MSL stage on the basis of (b). \textbf{(d)} \textit{w/o} DPI removes the dynamic persona identification on the basis of (c).

From the results, we observe that (1) DPOC plays a crucial role in enhancing the acquisition of better persona information, and the elimination of criterion samples significantly diminishes the model's effectiveness. This is because the model can pay more attention to persona-related tokens after deep personalized alignment. Relevant case study can be found in Appendix~\ref{appendix: case_study}. Additionally, the findings suggest that merely employing DPO falls short in substantially improving the overall performance of models. This is because the preference alignment of DPO is not optimized for problems that can arise from personalized dialogue generation task, as illustrated in $\S$ \ref{sec: data_construction}. Furthermore, the diminished effectiveness observed upon removing static and dynamic persona identifiers underscores the importance of reorganizing training data before the supervised fine-tuning process.

\subsubsection{Effect of Sessions}

\begin{figure}[!h]
\centerline{\includegraphics[width=0.5\textwidth]{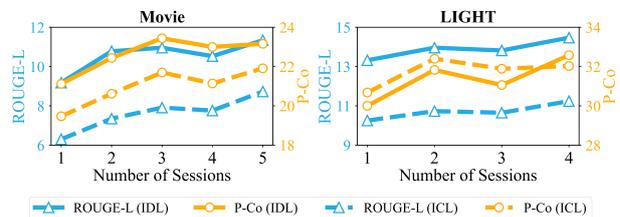}}
\caption{Experiments with different numbers of dialogue sessions on the Movie and LIGHT.}
\label{fig:sessions}
\end{figure}
In this work, we make the model learn personality-related information from the dialogue sessions and generate personalized responses. 
We present the performance of IDL and ICL under different demonstrations (dialogue sessions) to compare the learning efficiency of them.
Figure~\ref{fig:sessions} illustrates that similar to ICL, with the increase in the number of dialogue sessions, there is a general improvement in the quality of responses of IDL. 
However, as a specialized learning method for dialogue, IDL exhibits a faster learning ability under different dialogue sessions than ICL, indicating the effectiveness of our proposed mutual supervised learning and deep personalized alignment.
Benefits from these advancements, IDL paves a new road to develop and update dialogue systems in an online manner.


\section{Conclusion}
In this study, we introduce a framework In-Dialogue Learning (IDL) designed for personalized dialogue generation task. Unlike previous approaches, our framework directly derives persona information from dialogues without the need of pre-defined profiles and is widely applicable to LLMs. 
The efficacy of IDL in producing personalized responses is validated through both automatic and human evaluation results.

\clearpage
\section*{Limitations}
First, given the complexity of large-scale experiments, we limited our research to the more representative LLaMA-2 series models. This approach does not ensure favorable outcomes across all pre-trained large language models. Moreover, the capacity of IDL to manage highly diverse or conflicting persona traits within dialogue sessions has not been examined, which may restrict its use in situations involving non-coherent or changing user identities. Additionally, while the datasets employed in our study consistently includes personality information within dialogues, this may not hold true in real-world applications.

\section*{Ethics Statement}
Dialogues and persona information often contain sensitive information about individuals, which could result in breaches of privacy. We took measures to ensure that the datasets utilized in our experiments were strictly confined to the scope of the study and did not include any sensitive personal information.

The datasets employed in this research are publicly available, and the models we utilize adhere to their licenses, meeting both academic standards and ethical guidelines.

\section*{Acknowledgements}
This work was supported by the National Natural Science Foundation of China (NSFC Grant No. U21B2027, 62122089), Beijing Outstanding Young Scientist Program NO. BJJWZYJH012019100020098, and Intelligent Social Governance Platform, Major Innovation \& Planning Interdisciplinary Platform for the ``Double-First Class'' Initiative, Renmin University of China, the Fundamental Research Funds for the Central Universities, and the Research Funds of Renmin University of China.

\bibliography{ref}
\bibliographystyle{acl_natbib}

\appendix

\newpage
\section{Appendix}
\label{sec:appendix}
\subsection{Supplementary Results}
\label{appendix:supplement_exp}
We present the evaluation results in terms of BLEU-1/2/3/4 in Table \ref{tab:add_metrics}. The prompt used in LLaMA2-System is ``Here are your persona settings, your reply must be consistent with the persona: \{profile\}''. From the results, we can conclude that IDL holds consistent advantages over baseline methods on all BLEU metrics. 

\begin{table*}[htbp]
\centering

\begin{tabular}{ccccccccccc}
\toprule

\textbf{Dataset} & \textbf{Size} & \textbf{Model} & \textbf{BLEU-1} & \textbf{BLEU-2} & \textbf{BLEU-3} & \textbf{BLEU-4} \\
\midrule
\multirow{6}{*}{ConvAI2} & 124M & MSP & 8.19 & 2.34 & 1.56 & 0.73 \\
 & 125M & CLV & 11.85 & 4.29 & 2.41 & 1.02 \\
 & 7B & LLaMA-2 IDL & 52.40 & 25.43 & 12.74 & 8.40 \\
 & 7B & LLaMA-2 gold & 54.56 & 27.98 & 15.37 & 10.80 \\
 & 13B & LLaMA-2 IDL & 54.48 & 28.42 & 14.77 & 9.89 \\
 & 13B & LLaMA-2 gold & 55.32 & 28.71 & 16.00 & 11.33 \\
\midrule
\multirow{7}{*}{Movie} & 7B & Vicuna & 14.76 & 5.53 & 2.36 & 1.29 \\
 & 7B & LLaMA-2 ICL & 6.12 & 3.07 & 1.36 & 0.72 \\
 & 7B & LLaMA-2 IDL & 31.60 & 11.74 & 4.89 & 2.87 \\
 & 13B & Vicuna & 12.82 & 4.01 & 1.43 & 0.70 \\
 & 13B & WizardLM & 29.60 & 10.45 & 6.03 & 3.83 \\
 & 13B & LLaMA-2 ICL & 15.04 & 7.00 & 3.65 & 2.07 \\
 & 13B & LLaMA-2 IDL & 32.56 & 13.00 & 5.56 & 3.32 \\
\midrule
\multirow{7}{*}{LIGHT} & 7B & Vicuna & 36.07 & 17.37 & 7.42 & 3.83 \\
 & 7B & LLaMA-2 ICL & 15.41 & 8.92 & 3.99 & 2.02 \\
 & 7B & LLaMA-2 IDL & 46.32 & 22.01 & 9.63 & 5.27 \\
 & 13B & Vicuna & 19.68 & 8.87 & 3.53 & 1.75 \\
 & 13B & WizardLM & 44.59 & 21.45 & 10.30 & 5.53 \\
 & 13B & LLaMA-2 ICL & 24.31 & 13.47 & 6.00 & 3.08 \\
 & 13B & LLaMA-2 IDL & 49.69 & 24.64 & 10.90 & 6.03 \\
\bottomrule
\end{tabular}
\caption{Evaluation results w.r.t. BLEU 1-4.}
\label{tab:add_metrics}
\end{table*}

\subsection{Implementation Details}
\label{appendix: details}

\paragraph{In-Dialogue Learning.} In the Mutual Supervised Learning stage, the maximum cluster number \(c\) is set to 3 and the maximum number of neighbors \(k\) is set to 5. Besides, the scaling coefficient \(\lambda\) is set to 5. We use LoRA for training. The rank is set to 8 and the lora\_alpha is set to 8. We adopt AdamW as the optimizer. We set adam\_beta1, adam\_beta2 and adam\_epsilon to 0.9, 0.999 and \(1e^{-8}\), respectively. We use cosine schedule to warm up. The batch size is set to 4 and the learning rate is set to \(5e^{-5}\). We finetune our model on ConvAI2 dataset for 2 epochs. Each epoch takes around 40 minutes. The training of this process is completed on one Nvidia A100 GPU.

In the Deep Personalized Alignment stage, we set the penalty of DPOC to 2. The batch size is set to 1 and the learning rate is set to \(1e^{-5}\). We use LoRA for training. The rank is set to 8 and the lora\_alpha is set to 8. We adopt AdamW as the optimizer. We set adam\_beta1, adam\_beta2 and adam\_epsilon to 0.9, 0.999 and \(1e^{-8}\), respectively. We use cosine schedule to warm up. 

We fine-tune our models on the DPOC dataset, which is built based on ConvAI2. For each sample, we use the ground truth of ConvAI2 as the chosen sample, and select the one with the worst quality among the candidate responses (the candidate responses has been sorted by quality in the ConvAI2 provided by parlai) as the rejected sample. As for the criteria sample, we randomly select a type (line 386) and build it according to its corresponding construction method. The number of training epochs is set to 1. Each epoch takes around 12 hours. The training of this process is completed on one Nvidia A100 GPU.

\paragraph{Persona Extractor.} The original personaExt dataset contains 35K samples, with each sample containing a sentence and a triple <subject, relation, object>, which is a description of the persona information in the sentence. For example, the triple of the sentence "I have an apple, it is juicy." is <I, have, apple>. In order to adapt to the format of the conversation, we simply modified the original dataset by stacking multiple samples together to simulate the form of multiple sentences in a conversation. The modified dataset has 4K samples.

We select LLaMA-2 7B as the base model of Persona Extractor, and formalize the learning task as sequence-to-sequence generation (i.e., the model decodes the triples from a given sentence). Therefore, as for the input data, we concatenate sentences to form the input sequence, and use \textbackslash n as the separator. The form of output data is similar to the input data. We treat the triples corresponding to the statements as strings ``<{object}, {relation}, {subject}>'', and concatenate them to form the output sequence with \textbackslash n as the separator. The order of triples in the output sequence is the same as the order of their corresponding sentences in the input sequence.

We use 90\% of the data as the training set and the remaining 10\% as the validation set. To test the accuracy of the trained Persona Extractor, for each sample in validation set, we concatenate sentences to form the input sequence. After that, we can get the output sequence of the Persona Extractor. We use the regex ``<.*?,.*?,.*?>'' to parse the output sequence. If one element of the triple is different from the ground truth, then the sample is judged ``fals''. The accuracy of the Persona Extractor reached 87\% in validation.

\subsection{convED}
\label{appendix: convED}
Similar to Edit distance, convED also employs three operations: Insertion, Deletion, and Substitution. It calculates the shortest distance using Dynamic Programming (DP). However, unlike Edit distance, convED operates on sentences within dialogues, resulting in a distinct approach to distance calculation.

Assuming dialogue A comprises $m$ sentences and dialogue B comprises $n$ sentences, we obtain an $m \times n$ matrix $\text{lev}$, where $\text{lev}(i, j)$ represents the shortest edit distance between the first $i$ sentences of dialogue A and the first $j$ sentences of dialogue B. The costs of the three operations of convED are as follows:

\noindent \textbf{Insertion} Insert $B_j$ into dialogue A. The edit distance $\text{lev}_{ins}$ is updated as:

\begin{equation}
    \text{lev}_{ins}(i, j) = \text{lev}(i, j-1) + 1
\end{equation}

\noindent \textbf{Deletion} Delete $A_i$ from dialogue A. The edit distance $\text{lev}_{del}$ is updated as:

\begin{equation}
    \text{lev}_{del}(i, j) = \text{lev}(i-1, j) + 1
\end{equation}

\noindent \textbf{Substitution} Substitute sentence $A_i$ to align with $B_j$. The edit distance $\text{lev}_{sub}$ is updated as:

\begin{equation}
    \text{lev}_{sub}(i, j) = \text{lev}(i-1, j-1) + \lambda \cdot w_{sub}(A_i, B_j)
\end{equation}

\noindent The scale parameter $\lambda$ regulates the substitution cost, with both insertion and deletion costs being fixed at 1. $w_{sub}$ is a function that calculates the semantic similarity of two sentence vectors:

\begin{equation}
    w_{sub}(s_1, s_2) = 
    \begin{cases}
    \infty \ \ \ \  \text{if } r(s_1) \neq r(s_2) \\
    1 - \text{cos}(Enc(s_1), Enc(s_2))
    \end{cases}
\end{equation}

\noindent where $Enc$ is the encoder, used to encode sentences into vector space. It's important to highlight that sentences uttered by different individuals in a conversation, even if they share semantic similarities, cannot be aligned through substitution. Consequently, the function $r(*)$ is employed to identify the speaker of a sentence. Cosine similarity is then calculated for sentences from the same speaker, while the substitution cost between sentences from different speakers is considered infinite.

Finally, $\text{lev}(i, j)$ is the minimum cost of these three operations:

\begin{equation*}
    \text{lev}(i, j) = 
    \begin{cases} 
    \max(i,j) \ \ \ \ & \text{if } \min(i, j) = 0 \\
    \min \begin{cases}
    \text{lev}_{ins}(i, j)\\
    \text{lev}_{del}(i, j)\\
    \text{lev}_{sub}(i, j)
    \end{cases} & \text{otherwise}
    \end{cases}
\end{equation*}

\subsection{Completed Loss Function for DPOC}
\label{appendix: dpoc}
Please refer to Equation~\ref{eq: dpoc}.

\subsection{Case Study}
\label{appendix: case_study}

\begin{figure*}[]
\centerline{\includegraphics[width=0.9\textwidth]{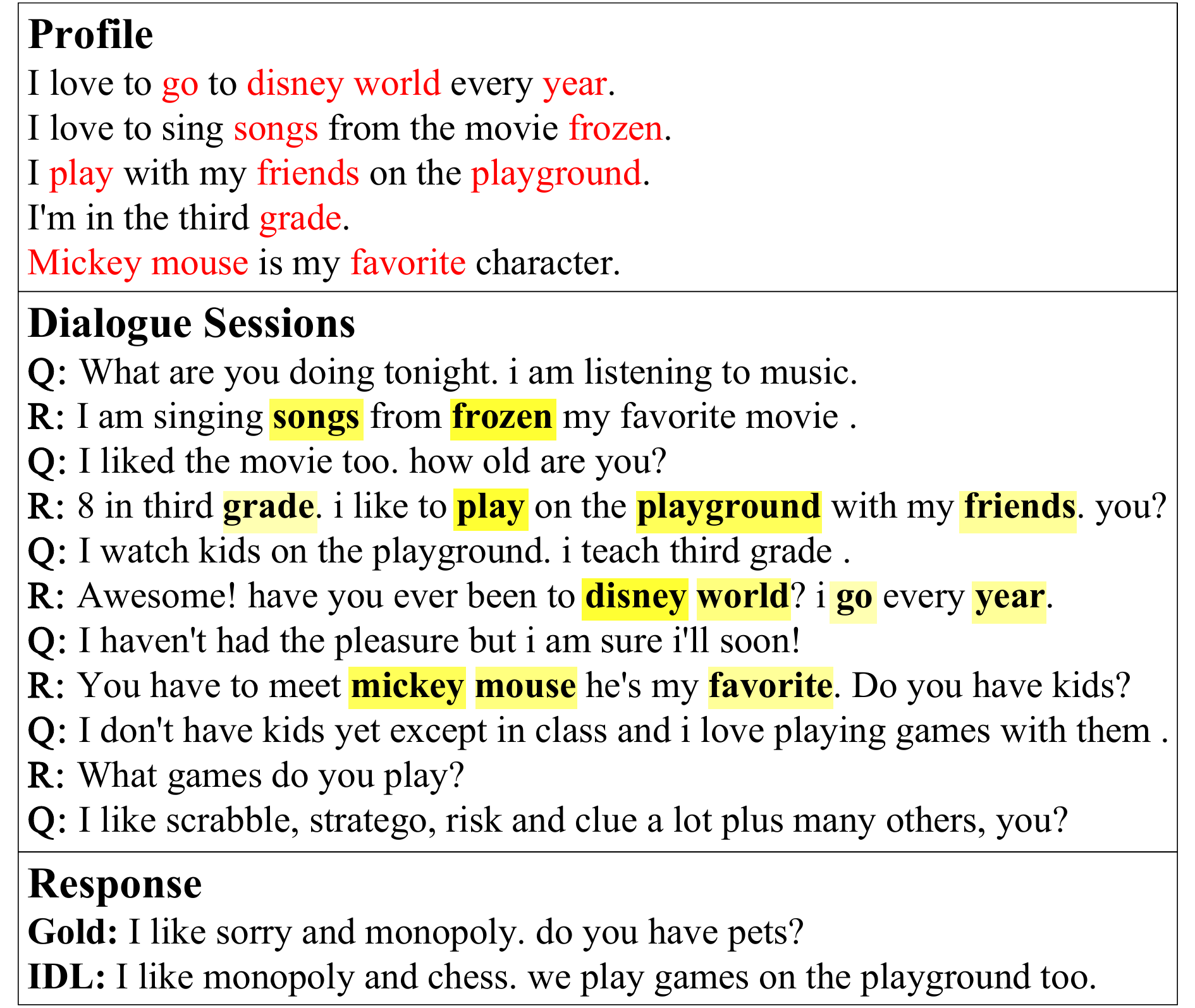}}
\caption{A case study. Keywords in the profile are marked in red, while the corresponding keywords that have high attention weight within dialogue sessions are bolded and highlighted with a yellow background.}
\label{fig:case_study}
\end{figure*}

To investigate the specific content within dialogue sessions that a model trained with IDL focuses on when crafting responses, we conducted an analysis of the attention weights during the reply generation process, as illustrated in Figure~\ref{fig:case_study}. We identified the top 100 tokens receiving the highest attention within the dialogue sessions and examined their correspondence with the personality-related keywords found in the gold profile. The experimental findings indicate that the LLaMA-2-13B-Chat model typically concentrates on an average of 9 keywords. However, the same model, once implemented with IDL, shows an enhanced focus on 13 keywords. This improvement suggests that IDL significantly enhances the model's ability to precisely leverage persona information within dialogues.

\subsection{Low-resource Scenarios}
We hope that the current conversation and its historical conversations are similar, so that the model can get more relevant information available from historical conversations. Therefore, in the Mutual Supervised Learning stage, we cluster input conversations so that they share more persona information (Static Persona Identification) and minimize the distance between the current conversation and historical conversations (Dynamic Persona Identification).

Of course, even so, we still can not guarantee that current conversation is similar to its historical conversations. Therefore, we add the following experiments. For each sample in original ConvAI2 test set, we replace the historical conversations with those in other clusters, so that the similarity between the historical conversations and the target conversation in this example is reduced. Results are shown in Table~\ref{table: low-resource}. The model used in this experiment is LLaMA-2 IDL 13B.

We can observe that the performance of the model has declined slightly. This is because the similarity between the historical conversations and the current conversation is the basic guarantee for IDL to be effective. When the current conversation involves a certain topic, IDL will focus on similar parts in historical conversations, thus completing the simulation of the persona. Therefore, when the similarity between the historical conversations and the current conversation decreases, the performance of IDL will also be affected.

\subsection{Generalizability}
To assess the generalizability of IDL, we also utilize LLaMA as the base model. Experimental results are presented in Table~\ref{table:llama-movie} and Table~\ref{table:llama-light}.

We repeated the training process of LLaMA-2 IDL using LLaMA and obtained LLaMA IDL. According to the results, LLaMA IDL surpasses Vicuna in multiple metrics, which further illustrates the effectiveness of IDL.

\begin{table*}[]
\centering
\tiny
\begin{tabular}{@{}clccccccccc@{}}
\toprule
\multicolumn{1}{l}{\textbf{Size}} & \textbf{Similarity} & \multicolumn{1}{l}{\textbf{BLEU-1}} & \multicolumn{1}{l}{\textbf{BLEU-2}} & \multicolumn{1}{l}{\textbf{BLEU-3}} & \multicolumn{1}{l}{\textbf{BLEU-4}} & \multicolumn{1}{l}{\textbf{ROUGE-L}} & \multicolumn{1}{l}{\textbf{Dist-1}} & \multicolumn{1}{l}{\textbf{Dist-2}} & \multicolumn{1}{l}{\textbf{Coh.}} & \multicolumn{1}{l}{\textbf{Coh-Con}} \\ \midrule
13B & Original & 54.48 & 28.42 & 14.77 & 9.89 & 20.05 & 87.78 & 97.45 & 98.48 & 19.63 \\
13B & Out-of-Cluster & 52.37 & 25.93 & 11.69 & 6.97 & 16.49 & 90.84 & 98.95 & 97.82 & 6.60 \\
7B & Original & 52.40 & 25.43 & 12.74 & 8.40 & 18.98 & 86.13 & 86.97 & 96.86 & 13.26 \\
7B & Out-of-Cluster & 51.45 & 24.53 & 11.56 & 6.84 & 16.47 & 87.70 & 97.99 & 95.27 & 7.15 \\ \bottomrule
\end{tabular}
\caption{Results for low-resource scenario.}
\label{table: low-resource}
\end{table*}

\begin{figure*}[h]
\begin{equation}
\label{eq: dpoc}
    \begin{aligned}
\mathcal{L} _{DPOC}(\pi _{\theta}; \pi _{ref}) = &-\mathbb{E} _{(x,y _w,y _r, y _l) \sim \mathcal{D}} [ \log \sigma ( \beta \log \frac{\pi _{\theta}(y _w | x)}{\pi _{ref}(y _w | x)} - \beta \log \frac{\pi _{\theta}(y _l | x)}{\pi _{ref}(y _l | x)} ) \\ 
&- \min (0, \lambda \log \frac{\pi _{\theta(y _w|x)}}{\pi _{ref}(y _w|x)} - \lambda \log \frac{\pi _{\theta(y _r|x)}}{\pi _{ref}(y _r|x)} ) \\
&- \min (0, \lambda \log \frac{\pi _{\theta(y _r|x)}}{\pi _{ref}(y _r|x)}- \lambda \log \frac{\pi _{\theta(y _l|x)}}{\pi _{ref}(y _l|x)} ) ],
\end{aligned}
\end{equation}
\end{figure*}

\begin{table*}[]
    \centering
    \begin{tabular}{cccccccc}
        \toprule
        \textbf{Model} & \textbf{BLEU-1} & \textbf{BLEU-2} & \textbf{ROUGE-L} & \textbf{Dist-1} & \textbf{Dist-2} & \textbf{P-F1} & \textbf{P-Co} \\
        \midrule
        Vicuna 7B & 14.76 & 5.33 & 5.44 & 71.45 & 63.58 & 11.13 & 17.05 \\
        LLAMA 7B ICL & 11.14 & 3.70 & 4.90 & 53.13 & 55.85 & 10.25 & 15.33 \\
        LLAMA 7B IDL & 22.55 & 9.01 & 8.79 & 59.58 & 70.53 & 20.63 & 20.11 \\ \midrule
        Vicuna 13B & 12.82 & 4.01 & 3.88 & 75.37 & 60.53 & 6.54 & 14.22 \\
        LLAMA 13B ICL & 11.27 & 3.74 & 4.42 & 45.85 & 46.49 & 8.83 & 14.85 \\
        LLAMA 13B IDL & 24.11 & 9.69 & 8.79 & 68.02 & 78.43 & 18.25 & 20.91 \\
        \bottomrule
    \end{tabular}
    \caption{Generalizability experiment results on Movie dataset}
    \label{table:llama-movie}
\end{table*}

\begin{table*}[]
    \centering
    \begin{tabular}{cccccccc}
        \toprule
        \textbf{Model} & \textbf{BLEU-1} & \textbf{BLEU-2} & \textbf{ROUGE-L} & \textbf{Dist-1} & \textbf{Dist-2} & \textbf{P-F1} & \textbf{P-Co} \\
        \midrule
        Vicuna 7B & 36.07 & 17.37 & 10.52 & 83.27 & 90.56 & 16.53 & 23.4 \\
        LLAMA 7B ICL & 19.39 & 7.80 & 6.83 & 61.36 & 64.83 & 10.75 & 17.01 \\
        LLAMA 7B IDL & 46.89 & 23.18 & 13.87 & 80.68 & 93.48 & 24.07 & 31.21 \\ \midrule
        Vicuna 13B & 19.68 & 8.87 & 5.87 & 59.85 & 58.07 & 8.27 & 16.11 \\
        LLAMA 13B ICL & 22.57 & 9.18 & 7.19 & 57.30 & 60.78 & 12.25 & 17.98 \\
        LLAMA 13B IDL & 48.77 & 24.25 & 13.99 & 83.42 & 95.70 & 23.52 & 31.75 \\
        \bottomrule
    \end{tabular}
    \caption{Generalizability experiment results on LIGHT dataset}
    \label{table:llama-light}
\end{table*}

\end{document}